# Applying Automated Machine Translation to Educational Video Courses

Linden Wang[*]

**Abstract**

We studied the capability of automated machine translation in the online video education space by automatically translating Khan Academy videos with state-of-the-art translation models and applying text-to-speech synthesis and audio/video synchronization to build engaging videos in target languages. We also analyzed and established two reliable translation confidence estimators based on round-trip translations in order to efficiently manage translation quality and reduce human translation effort. Finally, we developed a deployable system to deliver translated videos to end users and collect user corrections for iterative improvement.

---

[*]Linden Wang, linwan@ucdavis.edu, Department of Computer Science, University of California Davis, Davis, California, USA, (ORCID: 0000-0002-5758-4524)

# 1. Introduction

Online learning has received massive popularity in recent years. The amount of American students enrolled in online video courses has risen by 30% by 2018, and areas like Asia and the Middle East are already taking initiatives to improve and broaden educational video content (Palvia et al., 2018). Since the Coronavirus pandemic, curriculum worldwide has adapted, and teachers and students are starting to depend more and more on online educational videos (Nambiar, 2020).

With English being the global language of communication, and the rising availability of English educational video content, it is important for non-English speakers to have access to these resources. Access to open educational resources (OERs) is out of reach for non-English speakers in many developing countries, resulting in educational inequity due to the language barriers (Tahirsylaj et al., 2018; Ruipérez-Valiente et al., 2022; Godwin-Jones, 2014; Karakaya and Karakaya, 2020). There is a need for scalable methods to localize educational videos, yet current options, like auto-generated captions on platforms like YouTube, fail to make an engaging learning experience.

Several organizations have worked to localize educational video content through human translation. The company NetEase translated a subset of Khan Academy videos to Chinese and estimated more than 200 million page views over a four-year period, indicating a clear demand (Rao et al., 2017). Regional online video courses, which are frequently being added online, have also been found to attract a large and diverse population (Ruipérez-Valiente et al., 2022). Another study where Chilean students were given Spanish subtitles to Khan Academy videos resulted in improved math proficiency, further suggesting the benefits of translating educational video content (Light, 2016). In addition to benefits and demand for translated video courses, many such organizations have also faced challenges. The Teacher Education in Sub Saharan Africa (TESSA), a project aiming to localize OERs to improve teaching in Sub Saharan Africa, observed that significant resources were allocated to the project, which spanned more than a year to finalize. The expense associated with professional translations and their linguistic complexity, particularly for young students, was also recognized as a challenge (Wolfenden, Buckler, & Keraro, 2012). Khan Academy takes a different approach to localization, where volunteers around the world apply to translate (Khan Academy, 2020). These efforts require tedious manual labor, indicating the potential need of a scalable and automated process.

There have been previous efforts performing automated translation on online courses such as Khan Academy videos (Bendou, 2021), but the process is still not quite automatic and scalable. TraMOOC, a systematic way to apply machine translation in education and use multimodal evaluation techniques to ensure quality, has been proposed but requires a large amount of resources to control quality (Kordoni et al., 2016; Kordoni et al., 2015). Though some work has been done applying automated machine translation to educational video courses, many are still unconvinced on accuracy of translated content (Way, 2018).

In this paper, we aim to find a robust machine translation technique that is resource efficient for massive online learning content, while maintaining a high translation accuracy. Towards such a goal, we report in this paper: 1) a method to automatically translate Khan Academy videos; 2) two confidence estimators to evaluate translation quality and flag poorly translated sentences for human correction; 3) a system that delivers translated videos to end users via YouTube and the Khan Academy website, and also allows human correction of low-confidence translations via mass collaboration.

# 2. Methods

## 2.1 Technology Adoption

There are different approaches in automated video translation. Direct speech-to-speech translation still faces challenges in quality evaluation, delay time, and translation accuracy due to lack of context (Salesky et al., 2021; Dhawan, 2022). Automatic Speech Recognition (ASR) to convert audio to text, followed by text translation is a more common method of video translation (Alharbi et al., 2021; Chan & Wang, 2021). In this paper, we use the latter method combined with text-to-speech synthesis to generate audio, and audio/video synchronization to deliver educational video content in target languages.

## 2.2 Speech-To-Text Conversion

We began by collecting transcripts for videos using YouTube's transcription feature. This service utilizes automated speech recognition to transcribe videos, and encourages video authors to make edits and corrections. We found these transcripts to be reasonably accurate. The collected transcripts follow a consistent format, with sentences mapped to the timestamp in which the sentence was spoken in the video.

## 2.3 Text Translation to Target Languages

Our goal was to translate the sentences while preserving the mapping of each translated sentence to the original sentence timestamp to ensure the translated audio could be synchronized with the original video. This is necessary because different languages have different speech tempos. Initially, we considered translating each sentence, since the original transcript already mapped sentences to timestamps. However, this could lose the context provided by neighboring sentences, so for our final video translation processing algorithm, we combine the sentences in the transcript into a full text, translate the text, and tokenize it (split the text to the sentence level). Sentences were translated using two different neural machine translators - Google Translate and DeepL (DeepL, 2022) - into two languages - Chinese and Spanish. Then, we confirm that there is a one-to-one match between the original sentences and the translated sentences. Each translated sentence is mapped to an original sentence, and therefore mapped to a starting timestamp.

## 2.4 Text-To-Speech Conversion

We then performed text-to-speech synthesis on each translated sentence to build audio files. We used Google's text-to-speech service. Text-to-speech synthesis generally does not have issues in terms of accuracy and is not judged on the ability to correctly read words, instead it is primarily evaluated in terms of how accurately it can mimic human speech by not sounding robotic or unusual (Mendelson and Aylett, 2017). We recognize that more natural sounding text-to-speech systems have emerged recently in artificial intelligence (Kreuk et al., 2022), and that there is the possibility to experiment with those systems in the future to create more engaging educational videos.

## 2.5 Video Assembling and Synchronization

We then assembled the translated audio with the original video. Taking the translated audio fragments generated by text-to-speech synthesis, we inserted pauses to ensure the visual and audio components of the video were synchronized. If a translated sentence audio fragment exceeded the duration of the corresponding original sentence, we would take the video frame at the time when the original sentence was finished, and suspend that frame until the translated sentence finished. If the length of the translated audio fragment was shorter than the duration of the original sentence, a pause was inserted at the end of the translated audio fragment, allowing the video to synchronize.

However, not every video's transcript could undergo this matching process due to faulty tokenization, resulting in some loss of data. Occasionally, the tokenizer encountered difficulties in accurately segmenting the translated text into individual sentences. This resulted in instances where a tokenized segment was erroneously split into two separate sentences or an incomplete sentence. This hindered the accurate mapping of translated sentences back to their original counterparts. While potential remedies exist, such as testing different tokenizers and algorithms to separate the transcript text, and repeating the tokenization process for a predetermined number of iterations before considering video exclusion, these approaches were not incorporated within the confines of this paper.

# 3. Course Video Selection

Raw Khan Academy videos were selected and downloaded from YouTube. We chose videos from two popular subjects (reading and math), and grade levels (primary to high school/college). Specific categories are listed in Table 1.

**Table 1.** Selected Khan Academy Video Data for this study.

|                   | Number of Videos Downloaded | Number of Videos Used | Total Used Sentences |
|-------------------|-----------------------------|-----------------------|----------------------|
| **Grade 2 Math**  | 50                          | 42                    | 1506                 |
| **Grade 2 Reading** | 20                        | 10                    | 588                  |
| **Grade 4 Math**  | 171                         | 58                    | 2563                 |
| **Grade 4 Reading** | 26                        | 15                    | 968                  |
| **Grade 8 Math**  | 201                         | 42                    | 2542                 |
| **Grade 8 Reading** | 15                        | 9                     | 586                  |
| **Grade 9 Reading** | 12                        | 7                     | 409                  |
| **Early Math**    | 112                         | 29                    | 1041                 |
| **Linear Algebra**| 131                         | 20                    | 2366                 |
| **Precalculus**   | 268                         | 29                    | 1732                 |

## 4. Translation Confidence Determination

A crucial part of automated machine translation is to automatically determine if translations are accurate, therefore we established a translation confidence metric for our work. Many quality estimators have been proposed previously. For example, BLEU (Papineni, 2002) scores are a popular metric, but require reference text in target languages. Round-trip evaluation is a process of translating a translated sentence back to the original language and evaluating its similarity with the original sentence, not requiring reference text. Though initially deemed unreliable, it has been shown to work well with neural machine translators, and round-trip scores are independent of the type of translator (Moon et al., 2020). To analyze similarity of original sentences and back-translated sentences, we choose quality estimators sBERT (Reimers & Gurevych, 2019) and BERTScore (Zhang et al., 2019), which outperform other sentence similarity metrics (Moon et al., 2020). Each BERTScore score includes a precision, recall, and F1 value, and we use F1 in our study.

Our goal was to find some BERTScore or sBERT threshold based on comparison of the original versus back-translated sentences, which could determine the confidence of translation accuracy. To find this threshold, we took a sample of DeepL-translated, Chinese videos with 300 random sentences in reading (2nd grade reading, 4th grade reading, 9th grade reading), and 300 random sentences in math (early math, 4th grade math, linear algebra). For each sentence, we manually evaluated if the original and translated sentences were truly equivalent, and if the original and back-translated sentences were truly equivalent. These evaluation results served as our groundtruth.

We then tested whether the original versus back-translated sentence comparison can serve as a predictor for translation confidence. Using the original versus translated sentence evaluation as ground truth, we took two sentences with equivalent meanings (a match) as a positive, and two sentences with significantly different meanings as a negative.

The results comparing the prediction to ground truth (Table 2) showed that the original versus back-translated sentence comparison can predict the translation accuracy with a high confidence. The reported percentages are relative to the numbers of all the sentences in each subject. For example, the false positive percentage for reading is the proportion of false positives in the 300 reading sentences we evaluated. For each subject, the false positive percentage is below 3% while the false negative percentage is below 6%. We observed that math videos are more likely to produce false positive and false negative sentences. A possible reason is math sentences have less context clues and can sometimes be unpredictable. For example, one original sentence in a math video, "We'd put the seven in the ones place.", had a back translation of "We'll put seven people in one place.", indicating a clear misunderstanding of the sentence. Second, across both subjects, we also noticed a higher false negative rate which could be partly attributed to homonym confusion. When a sentence is translated poorly (for example, by using the wrong meaning of a homonym), the back-translation is more likely to also contain that incorrect interpretation because back-translations are independent of forward-translations. It is unlikely for the back-translation to take a translated sentence with the wrong homonym and produce a back-translated sentence with the correct homonym (causing a false positive).

Table 2. False Positive and False Negative Percentages (Taking Original vs. Translated as Truth and Original vs. Back-translated as Prediction), based on manually evaluated English/Chinese sentence pairs (300 from reading and 300 from math). False positive/false negative percentages refer to the proportion of sentences in that subject that were false positive/false negative.

|  | Reading | Math |
| --- | --- | --- |
| **False Positives** | 0.0% | 2.9% |
| **False Negatives** | 2.3% | 5.9% |

We then compared sentence-level BERTScore F1 and sBERT scores to the manual evaluations of the original versus back-translated sentences as the ground truth. Again we take two sentences with equivalent meanings (a match) as a positive, and two sentences with significantly different meanings as a negative. For example, a false positive represents a sentence pair which was manually determined to be unequivalent, but classified as equivalent by a BERTScore or sBERT threshold. If an original sentence does not match the back-translated sentence, but the calculated BERTScore score was 0.91 (assuming a BERTScore threshold of 0.90), then this would be a false positive. For each score, we experiment with different sBERT/BERTScore thresholds and count the number of false positive, false negative, and true positive sentences. We report the thresholds which result in a 1%, 2%, and 3% false positive percentage, along with the false negative and true positive percentages corresponding to each threshold (Table 3). Each percentage represents proportion relative to all the sentences in a specific subject.

Table 3. BERTScore F1 and sBERT Thresholds vs. True Positive/False Positive/False Negative Percentages, based on manually evaluated English/Chinese sentence pairs (300 from reading and 300 from math). Each percentage is relative to all the sentences in the respective subject.

|  | BERTScore F1 | True Positives | False Positives | False Negatives | sBERT | True Positives | False Positives | False Negatives |
| --- | --- | --- | --- | --- | --- | --- | --- | --- |
| **Reading** | 0.940 | 85.4% | 3% | 7.94% | 0.835 | 83.8% | 3% | 9.60% |
|  | 0.955 | 71.2% | 2% | 22.2% | 0.890 | 76.2% | 2% | 17.2% |
|  | 0.962 | 63.6% | 1% | 29.8% | 0.940 | 63.9% | 1% | 29.5% |
| **Math** | 0.956 | 63.0% | 3% | 24.6% | 0.881 | 63.9% | 3% | 23.6% |
|  | 0.959 | 58.7% | 2% | 28.9% | 0.931 | 51.1% | 2% | 36.4% |
|  | 0.965 | 49.5% | 1% | 38.0% | 0.947 | 45.2% | 1% | 42.3% |

Initially, for each threshold value we calculated a recall, precision, and F1 score, trying to optimize F1. For example, a BERTScore threshold of 0.907 results in a maximum F1 value of 0.938, corresponding to a 11% false positive percentage and 0.33% false negative percentage for DeepL chinese math videos. However, this is not a practical confidence measure for our application, as threshold scores will be used to flag potentially incorrect sentences. Many false positives will result in many incorrectly translated sentences not being flagged and therefore more likely to be skipped in the human correction step. On the other hand, many false negatives mean that human contributors need to review more translations and mark them as correct in our contribution system. This is more manageable without sacrificing quality, as during the review process, these false negative sentences can be quickly confirmed as correct translations by contributors with low effort. Therefore, we aimed to find a threshold to minimize false positives (as the top priority) without incurring too many false negatives.

We found a false positive percentage of 2-3% to be optimal. Further reducing it to below 2% would significantly increase the false negative percentage. From Table 3, we can also derive the percentage of false positives in all perceived positives. A false positive percentage of 2% corresponds to a false discovery rate (percentage of false positives in all perceived positives) of ≤2.7% for reading and ≤3.8% for math. This suggests that the quality estimators can be effective in filtering out most false positives in sentences regarded as being accurately translated. As shown in Table 3, a 0.955 BERTScore F1 or 0.890 sBERT threshold for reading (and a 0.959 BERTScore F1 or 0.931 sBERT threshold for math) represents the cutoff between accurate and inaccurate translations while maintaining a 2% false positive percentage.

We noticed a stricter BERTScore F1 and sBERT threshold is needed for math sentences, which is likely due to the unpredictability of math sentences as discussed earlier. Across reading and math videos, we see the percentages of false negative sentences are comparable between BERTScore F1 and sBERT, with BERTScore slightly outperforming sBERT, suggesting both to be effective metrics for determining the translation confidence. We also experimented with a combination of BERTScore F1 and sBERT thresholds to further reduce false negative rates but found no improvement.

Overall, the original vs. back-translated sentence similarity accurately represents translation quality, suggesting the confidence estimator of BERTScore and sBERT using back-translated sentences is practical. We used the 2% false positive percentage and corresponding thresholds (0.955 BERTScore F1 for reading and a 0.959 BERTScore F1 for math) for the rest of the experiment and implementation.

## 5. Translation Results and Analysis

In this section, we study performance of machine translation and its dependencies on translation model, target language, subject, and grade level. We then discuss how well machine translation handles challenging cases.

We took each combination of translation model, target language, and subject (for example: DeepL, Spanish, Reading) and computed the BERTScore, sBERT, and correct translation percentage. Correct Translation Percentage refers to the percentage of sentences in each category that exceeded a BERTScore F1 threshold of 0.955 (corresponding to a confidence level of 2% false positive percentage, as discussed in the previous section). Figure 1 shows variations of median BERTScore F1 and Correct Translation Percentage for the selected grade levels in both subjects (reading and math) for the two language models (Google Translate and DeepL).

**Figure 1.** Median BERTScore F1 and Correct Translation Percentage Trends across Reading and Math Videos

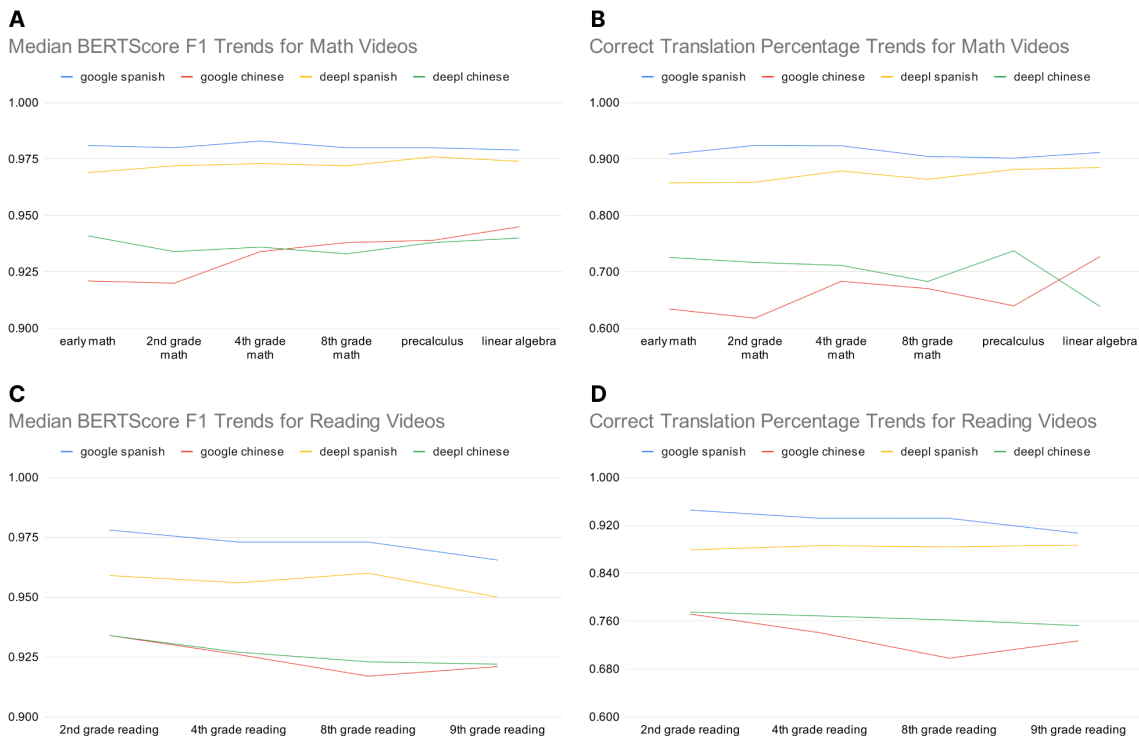

Figure 1 demonstrates that both BERTScore values and correct translation percentages are fairly consistent across the language models, target languages, subjects and grade levels. The largest factor affecting translation performance differences is the target language. The differences caused by different target languages are more prominent than differences due to 1) different translation models, 2) different subjects and 3) different grade levels. For example, Spanish appears to lead to more accurate translations than Chinese, regardless of whether Google Translate or DeepL is used. With the same target language, the performance is also consistent across language models, subjects and grade levels. For example, for Spanish, Google Translator consistently scores higher than DeepL, whereas for Chinese, DeepL scores slightly higher than or comparable to Google Translate.

Although we showed quality estimator variations using only BERTScore in Figure 1, sBERT score likely can provide similar results. As an example, we show detailed statistics of BERTScore, sBERT and correct translation percentage across grade levels, for reading videos translated into Spanish using DeepL (Table 4). The mean, median and StdDev of both BERTScore and sBERT are consistent at different grade levels, with BERTScore having smaller variation than sBERT scores. Table 4 also suggests that the choice of median over mean BERTScore in Figure 1 is insignificant, as mean BERTScore should show the similar variation patterns.

**Table 4**. DeepL Spanish Reading Video Statistics

|  | Mean BERTScore F1 | Median BERTScore F1 | StdDev BERTScore F1 | Mean sBERT | Median sBERT | StdDev sBERT | Correct Translation Percentage[1] |
|---|---|---|---|---|---|---|---|
| **Grade 2 Reading** | 0.974 | 0.982 | 0.021 | 0.910 | 0.959 | 0.132 | 0.879 |
| **Grade 4 Reading** | 0.977 | 0.981 | 0.020 | 0.914 | 0.956 | 0.116 | 0.886 |
| **Grade 8 Reading** | 0.978 | 0.981 | 0.018 | 0.920 | 0.960 | 0.107 | 0.884 |
| **Grade 9 Reading** | 0.970 | 0.979 | 0.017 | 0.915 | 0.950 | 0.107 | 0.887 |

[1] Correct Translation Percentage refers to the percentage of sentences in each category that exceeded a BERTScore F1 threshold of 0.955 (corresponding to a confidence level of 2% false positive percentage).

As grade levels increase, translation accuracy slowly declines for reading videos (Figure 1C and Figure 1D). This met our expectations, as the material should become more difficult to translate with increasing grade levels. While manually analyzing the reading translations, we note that there are special cases where lower grade level content results in lower translation accuracy. Khan Academy had several videos at the second grade reading level with lower scoring translations, because they contained more colloquial terms and deviating content in attempts to engage a younger audience. A particular low-scoring set of videos teaching word roots featured an exercise where the instructor was creating imaginary words. However, because videos were the minority, this was not reflected in the overall translation performance across grade levels.

We also note that in math videos translated to Chinese, Google Translate shows a sharp uptick in translation accuracy compared to DeepL for the linear algebra category. This suggests Google Translate to be a potentially stronger translation model in this specific case. It may be possible to improve translations by switching to the stronger translation model depending on grade level, target language, and subject.

Lastly, we look to understand how current language models handle common translation pitfalls, which can further provide insight on the feasibility of applying machine translation to educational videos. We first tested the common culprits associated with unreliable translations: homonyms, interchangeable words, and speaker mistakes (Al Sharou & Specia, 2022). Homonyms occur frequently in math videos, where, for example, place values can be confused with quantities. The sentence "We put seven in the ones place" can be correctly interpreted as putting seven in the ones digit of a number or incorrectly interpreted as putting a seven in one (singular) place. We notice the translation models handle these situations quite well although there are a few occurrences where homonyms are mistranslated. Interchangeable words, such as "magnitude" and "size" and "perpendicular" and "normal" in linear algebra videos, were no problem for either translator. In the case of speaker mistakes, where the speaker would correct themselves mid-sentence leaving several extraneous words in the transcript, both translators correctly ignored the extraneous words, with sBERT scores dropping slightly more.

Overall, we demonstrated that text-to-text translation can be a reliable method for our selected videos while maintaining a high translation accuracy. As Figure 1 shows, in the case of Spanish translation, we can achieve a correct translation percentage of about 86%–95%. For Chinese translation, the correct translation percentage is lower (62%–77%). Quality estimators BERTScore and sBERT can be effectively used for maintaining a desired confidence level (a low false positive percentage and a manageable false negative percentage) during translation without human involvement.

## 6. Implementation

To apply our findings to the real world educational space, we developed a system to automatically translate Khan Academy videos, and deliver the translated videos to end users and receive user contributions, based on analysis and findings in the previous sections. The system is composed of a chrome extension, a backend system, and a video processing module. Their interaction is shown in Figure 2.

**Figure 2**. Video Translation Application System Design

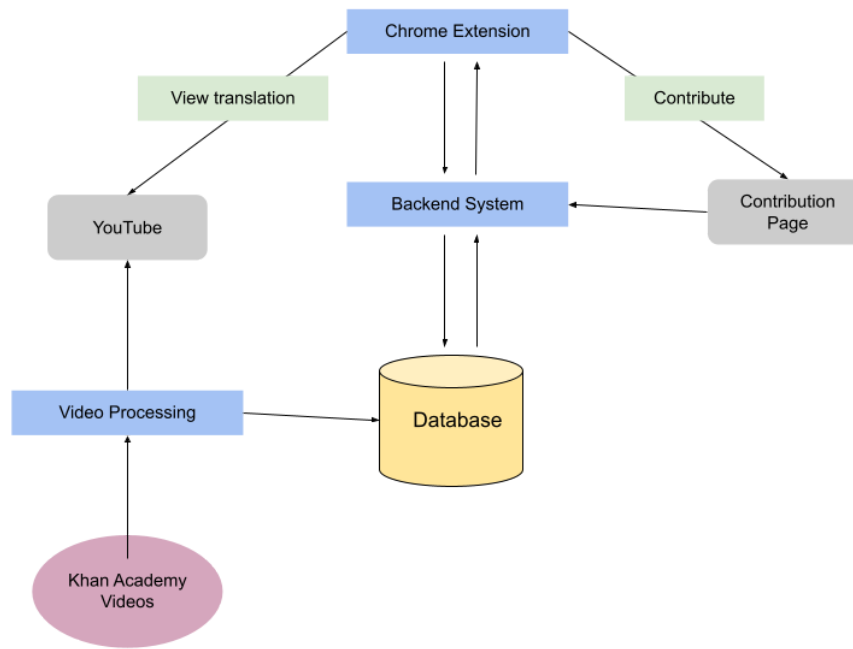

This user contribution option is based on crowdsourcing, and can allow efficient correction of machine translation inaccuracies. As other studies showed, crowdsourcing and automated quality control can result in professional translations from non-professional translators (Zaidan & Callison-Burch, 2011).

In most common use cases of our system, a user first finds a Khan Academy video on the Khan Academy website or YouTube. Using the chrome extension (Wang, 2022a), they have the option to view the translated video which is hosted on YouTube (Wang, 2022b).

The workflow of contribution is as follows: the user will be redirected from their video to the contribution page, authenticate themselves, view the sentences in the video, input their correction to inaccurately translated sentences, and submit. Sentences likely to be incorrect are flagged based on the quality confidence threshold proposed earlier, and highlighted for easier identification.

Periodically, a crawler evaluates proposed sentence contributions using the BERTScore and sBERT round-trip metric, and if these scores surpass previous ones, a new video is recompiled. With each iteration, videos are re-assembled with the more accurate translations to replace older versions. In the future, we look to analyze the improvement in educational video translation quality after we have accumulated sufficient human contributions.

## 7. Conclusion

We present a comprehensive approach to automated translation of educational video content at scale. For our research, we selected math and reading Khan Academy videos to translate to Spanish and Chinese. Our method uses automated speech recognition, text-to-text machine translation, text-to-speech synthesis, automated synchronization of original video and translated audio, and crowdsourcing to correct inaccurate translations. In the automated process, the text-to-text translation step can introduce errors, therefore, we also introduced two translation confidence estimators to identify inaccurate translations and flag them for user review. These estimators evaluate round-trip sentence similarity using BERTScore and sBERT. Finally, we deployed an online contribution system for users to correct flagged, inaccurately-translated sentences.

Overall, we find this approach to educational video translation to be effective in addressing the challenges of scaling the localization of educational video content, with benefits in its versatility and cost-effectiveness in translating education videos of diverse categories. Our translation confidence estimator does not require reference text, allowing efficient detection of poor translations to save human labor. The proposed translation contribution system allows human contributors to iteratively correct occasional machine translation errors and improve translation quality with small effort.

Future work may include extending the technique to other subjects, grade levels and target languages, further increasing the effectiveness of the confidence estimator, growing the contributing user base to enhance quality of translated videos, and evaluating translation quality before and after contribution iterations.

# References

Alharbi, S., Alrazgan, M., Alrashed, A., Alnomasi, T., Almojel, R., Alharbi, R., ... & Almojil, M. (2021). Automatic speech recognition: Systematic literature review. IEEE Access, 9, 131858-131876. https://doi.org/10.1109/ACCESS.2021.3112535

Al Sharou, K., & Specia, L. (2022). A Taxonomy and Study of Critical Errors in Machine Translation. In Proceedings of the 23rd Annual Conference of the European Association for Machine Translation (pp. 171-180). Ghent, Belgium: European Association for Machine Translation.

Bendou, I. (2021). Automatic Arabic Translation of English Educational Content Online using Neural Machine Translation: the Case of Khan Academy (Doctoral dissertation, Carnegie Mellon University). https://doi.org/10.1184/R1/16725304.v1

Chan, J. Y. & Wang, H. H. (2021). Speech Recorder and Translator using Google Cloud Speech-to-Text and Translation. Journal of IT in Asia, 9(1), 11-28. https://doi.org/10.33736/jita.2815.2021

DeepL. (2022). DeepL Translator [Software]. Retrieved from https://www.deepl.com/

Dhawan, S. (2022). Speech to Speech Translation: Challenges and Future. International Journal of Computer Applications Technology and Research, 11(03), 36–55. https://doi.org/10.7753/ijcatr1103.1001

Godwin-Jones, R. (2014). Global reach and local practice: The promise of MOOCS. Language Learning & Technology, 18(3), 5-15. https://doi.org/10125/44377

Karakaya, K., & Karakaya, O. (2020). Framing the Role of English in OER from a Social Justice Perspective: A Critical Lens on the (Dis)empowerment of Non-English Speaking Communities. Asian Journal of Distance Education, 15(2), 175-190. Retrieved from http://www.asianjde.com/ojs/index.php/AsianJDE/article/view/508

Khan Academy. (2020). Contribute [Web page]. Retrieved from https://www.khanacademy.org/contribute

Kordoni, V., Cholakov, K., Egg, M., Way, A., Birch, L., Kermanidis, K. L., ... & Orlic, D. (2015). TraMOOC: Translation for Massive Open Online Courses. In Proceedings of the 18th Annual Conference of the European Association for Machine Translation. https://aclanthology.org/W15-4935/

Kordoni, V., Van den Bosch, A., Kermanidis, K. L., Sosoni, V., Cholakov, K., Hendrickx, I., ... & Way, A. (2016, May). Enhancing access to online education: Quality machine translation of MOOC content. In Proceedings of the Tenth International Conference on Language Resources and Evaluation (LREC'16) (pp. 16-22). https://aclanthology.org/L16-1003/

Kreuk, F., Synnaeve, G., Polyak, A., Singer, U., Défossez, A., Copet, J., ... & Adi, Y. (2022). Audiogen: Textually guided audio generation. arXiv preprint arXiv:2209.15352.

Light, Daniel. (2016). Increasing Student Engagement in Math: The Study of Khan Academy Program in Chile. International Conference on Education, Research and Innovation (ICERI2016 Proceedings) (pp. 4593).

Mendelson, J., & Aylett, M. P. (2017). Beyond the Listening Test: An Interactive Approach to TTS Evaluation. Interspeech 2017. https://doi.org/10.21437/interspeech.2017-1438

Moon, J., Cho, H., & Park, E. L. (2020). Revisiting round-trip translation for quality estimation. European Association for Machine Translation. https://doi.org/10.48550/arXiv.2004.13937

Nambiar, D. (2020). The impact of online learning during COVID-19: students' and teachers' perspective. The International Journal of Indian Psychology, 8(2), 783-793. https://doi.org/10.25215/0802.094

Palvia, S., Aeron, P., Gupta, P., Mahapatra, D., Parida, R., Rosner, R., & Sindhi, S. (2018). Online education: Worldwide status, challenges, trends, and implications. Journal of Global Information Technology Management, 21(4), 233-241. https://doi.org/10.1080/1097198X.2018.1542262

Papineni, K., Roukos, S., Ward, T., & Zhu, W. J. (2002, July). Bleu: a method for automatic evaluation of machine translation. In Proceedings of the 40th annual meeting of the Association for Computational Linguistics (pp. 311-318). https://doi.org/10.3115/1073083.1073135

Rao, A., Hilton III, J., & Harper, S. (2017). Khan Academy videos in Chinese: A case study in OER revision. The International Review of Research in Open and Distributed Learning, 18(5). https://doi.org/10.19173/irrodl.v18i5.3086

Reimers, N., & Gurevych, I. (2019). Sentence-bert: Sentence embeddings using siamese bert-networks. The 2019 Conference on Empirical Methods in Natural Language Processing. https://doi.org/10.48550/arXiv.1908.10084

Ruipérez-Valiente, J. A., Staubitz, T., Jenner, M., Halawa, S., Zhang, J., Despujol, I., ... & Reich, J. (2022). Large scale analytics of global and regional MOOC providers: Differences in learners' demographics, preferences, and perceptions. Computers & Education, 180, 104426. https://doi.org/10.1016/j.compedu.2021.104426

Salesky, E., Mäder, J., & Klinger, S. (2021, December). Assessing Evaluation Metrics for Speech-to-Speech Translation. In 2021 IEEE Automatic Speech Recognition and Understanding Workshop (ASRU) (pp. 733-740). IEEE. https://doi.org/10.48550/arXiv.2110.13877

Tahirsylaj, A., Mann, B., & Matson, J. (2018). Teaching creativity at scale: Overcoming language barriers in a MOOC. International Journal of Innovation, Creativity and Change, 4(2), 1-19. https://www.ijicc.net/images/vol4iss2/Tahirsylaj_et_al.pdf

Wang, Linden. (2022a). Khan Academy Video Translator [Software]. https://chrome.google.com/webstore/detail/khan-academy-video-transl/gbpgbjnhccemhkjedfadjbekpmaoembh

Wang, Linden. (2022b). Khan Academy Videos Translated [YouTube Channel]. https://www.youtube.com/@KhanAcademyVideosTranslated/

Way, A. (2018). Quality expectations of machine translation. In Translation quality assessment (pp. 159-178). Springer, Cham. https://doi.org/10.48550/arXiv.1803.08409

Wolfenden, F., Buckler, A., & Keraro, F. (2012). OER Adaptation and Reuse across Cultural Contexts in Sub-Saharan Africa: Lessons from TESSA (Teacher Education in Sub-Saharan Africa). Journal of Interactive Media in Education, 16.

Zaidan, O. F., & Callison-Burch, C. (2011). Crowdsourcing Translation: Professional Quality from Non-Professionals. In Proceedings of the 49th Annual Meeting of the Association for Computational Linguistics: Human Language Technologies (pp. 1220-1229). Portland, Oregon, USA: Association for Computational Linguistics.

Zhang, T., Kishore, V., Wu, F., Weinberger, K. Q., & Artzi, Y. (2019). Bertscore: Evaluating text generation with bert. International Conference on Learning Representations. https://doi.org/10.48550/arXiv.1904.09675